\documentclass[10pt,twocolumn,letterpaper]{article}

\usepackage[accsupp]{axessibility} 
\usepackage{iccv}
\usepackage{times}
\usepackage{epsfig}
\usepackage{graphicx}
\usepackage{amsmath}
\usepackage{amssymb}

\usepackage{booktabs}
\usepackage{multirow}
\RequirePackage[format=plain,labelformat=simple,labelsep=period,font=small,compatibility=false]{caption}
\RequirePackage[font=footnotesize,skip=3pt,subrefformat=parens]{subcaption}
\usepackage{fontawesome5}
\def\x{$\times$}
\usepackage{pifont}
\newcommand{\cmark}{\ding{51}}%
\usepackage{xcolor}         
\definecolor{demphcolor}{RGB}{144,144,144}
\newcommand{\demph}[1]{\textcolor{demphcolor}{#1}}
\usepackage{colortbl}
\definecolor{baselinecolor}{gray}{.9}
\newcommand{\base}[1]{\cellcolor{baselinecolor}{#1}}
\newlength\savewidth\newcommand\shline{\noalign{\global\savewidth\arrayrulewidth
  \global\arrayrulewidth 1pt}\hline\noalign{\global\arrayrulewidth\savewidth}}

\usepackage{arydshln}

\usepackage{colortbl}
  
\usepackage[pagebackref=true,breaklinks=true,letterpaper=true,colorlinks,bookmarks=false]{hyperref}

\usepackage[capitalize]{cleveref}
\crefname{section}{Sec.}{Secs.}
\Crefname{section}{Section}{Sections}
\Crefname{table}{Table}{Tables}
\crefname{table}{Tab.}{Tabs.}  

\iccvfinalcopy 

\def\iccvPaperID{} 

\begin{document}

\title{What Can Simple Arithmetic Operations Do for Temporal Modeling?}

\author{%
Wenhao Wu$^{1,2}$\quad
Yuxin Song$^{2}$\quad
Zhun Sun$^{2}$\quad
Jingdong Wang$^{2}$\quad
Chang Xu$^{1}$\quad
Wanli Ouyang$^{3,1}$\\
$^1$The University of Sydney \qquad $^2$Baidu Inc.  \qquad $^3$Shanghai AI Laboratory\\
{\tt\small wenhao.wu@sydney.edu.au ~ \url{https://github.com/whwu95/ATM}}
}

\maketitle

\begin{abstract}
Temporal modeling plays a crucial role in understanding video content. 
To tackle this problem, previous studies built complicated temporal relations through time sequence thanks to the development of computationally powerful devices.
In this work, we explore the potential of four simple arithmetic operations for temporal modeling. Specifically, we first capture auxiliary temporal cues by computing addition, subtraction, multiplication, and division between pairs of extracted frame features. Then, we extract corresponding features from these cues to benefit the original temporal-irrespective domain. 
We term such a simple pipeline as an \textbf{A}rithmetic \textbf{T}emporal \textbf{M}odule (ATM), which operates on the stem of a visual backbone with a plug-and-play style.
We conduct comprehensive ablation studies on the instantiation of ATMs and demonstrate that this module provides powerful temporal modeling capability at a low computational cost. Moreover, the ATM is compatible with both CNNs- and ViTs-based architectures. Our results show that ATM achieves superior performance over several popular video benchmarks.
Specifically, on Something-Something V1, V2 and Kinetics-400, we reach top-1 accuracy of 65.6\%, 74.6\%, and 89.4\% respectively. The code is available at \url{https://github.com/whwu95/ATM}.

\end{abstract}

\section{Introduction}

\begin{figure}
    \centering
    \includegraphics[width=1\linewidth]{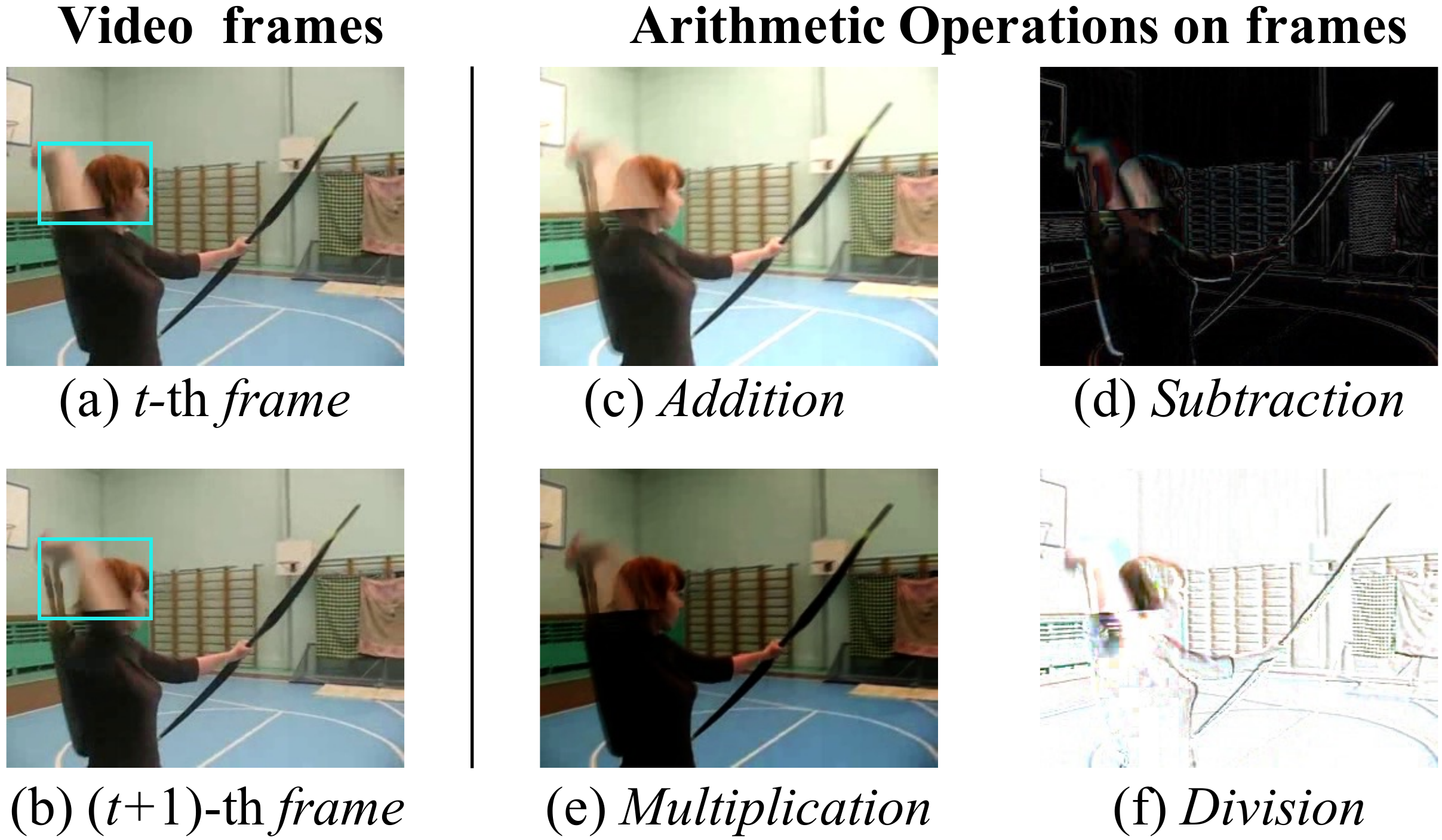}
    \caption{(a),(b): A pair of consecutive video frames where a cyan rectangle highlights the area around a moving hand. (c)-(f): Visualization of arithmetic operations (\faPlus, \faMinus, \faTimes, \faDivide) on consecutive frames. Zoom in for the best view.}
    \label{fig:teaser}
\end{figure}

\label{sec:intro}
Owing to the rapid development of the Internet and mobile devices, researchers can now work with massive amounts of video data. The time-sequenced data provides substantial information in understanding visual tasks such as human activity recognition. While significant progress has been made in extracting the semantics from a single image, how to model the temporal information effectively becomes an essential problem for video recognition.

Historically, temporal information has been constructed either by utilizing low-level streams (\eg optical flow, motion vector)~\cite{two-stream-stresnet,two-stream,tsn,mv-cnn}, or by building segregated time sequence with recurrent networks using features extracted from an image backbone~\cite{LRCN,li2017temporal,yue2015beyond}. After that, more works present 3D CNNs~\cite{i3d,c3d} and factorized 2D+1D CNNs~\cite{p3d,r2+1d,s3d} as the natural evolution from their 2D counterparts for dealing with 3D volumetric video data directly, thanks to the rapid improvement of computational capacity. Along this research line, many recent pieces of research follow the ``2D backbone + temporal interaction'' paradigm to develop efficient temporal modules~\cite{stnet,kwon2020motionsqueeze,selfy,li2020smallbignet,li2020tea,tsm,teinet,liu2021tam,off,tdn,wu2020MVFNet,dsanet,eco}. The strong point of this approach is that, the pre-trained image backbone can serve as a reasonable initialization, reducing training time and enabling the optimization of the entire video architecture in an end-to-end manner. Under this paradigm, the ways of temporal modeling can be roughly divided into two categories based on the type of interaction between frames: i) sequence-wise modeling (\eg, temporal convolution~\cite{slowfast,tsm,nonlocal,wu2020MVFNet}, dynamic convolution~\cite{liu2021tam,dsanet}, and temporal transformer~\cite{arnab2021vivit,timesformer,videoswin}); ii) pair-wise modeling (\eg, pseudo optical flow estimation~\cite{off}, similarity estimation~\cite{kwon2020motionsqueeze,selfy,corrnet}, \etc). 

We bring the latter into focus. Precisely, pair-wise modeling employs the relationship between an anchor frame and the others. Such relationships are considered to contain expert knowledge. For instance, a few previous works~\cite{kwon2020motionsqueeze,selfy,corrnet} have used correlation operations to estimate the similarity between two frames, while another work~\cite{off} has combined subtraction and Sobel operations to estimate optical flow. Despite the effectiveness of these limited attempts on CNN backbones~\cite{resnet}, the pair-wise modeling has not received enough attention and exploration when compared to sequence-wise modeling, especially in the current era of the rising popularity of vision transformer backbones~\cite{ViT,liu2021swin}. In this study, we aim to fill this gap by revisiting the pair-wise relationship with the most basic arithmetic operations: \emph{Addition} (\faPlus), \emph{Subtraction} (\faMinus), \emph{Multiplication} (\faTimes), and \emph{Division} (\faDivide), without extensive elaborate design. 

The arithmetic operations are commonly employed in ordinary vision and time series tasks, and their outputs have explainable physical meaning. Briefly, addition is used to implement integral to obtain  properties such as the average intensity of an image~\cite{derpanis2007integral} or the accumulated status of a feature sequence; On the contrary, subtraction detects changes over time and approximates the tendency of the motion such as optical flow~\cite{off}; Next, multiplication can be considered as a measure of similarity or correlation between (regions of) vectorized features of frames~\cite{kumar2005correlation}, such that the features that are unchanged along time shall be highlighted; Finally, on the opposite of multiplication, the division between frames locates the features that are changed intensely~\cite{vazquez2008new}. 
For easier understanding, we provide an intuition of the physical meaning of these operations using the original frames, as depicted in Figure~\ref{fig:teaser}. 
Notably, our work focuses on exploring these arithmetic operations on \textit{encoded feature representations}, rather than on raw frames.

In this work, we raise a question: 
\emph{Can simple arithmetic operations be employed for temporal modeling?}
To answer this question, we first abstract a simple video framework that divides the neural network into a temporal-irrespective stem and a temporal-interactive block. This enables the implementation of the temporal-irrespective stem using existing vision backbones, such as CNNs or Vision Transformers. We then propose an \textbf{A}rithmetic \textbf{T}emporal \textbf{M}odule (ATM), which conducts arithmetic operations between pairs of frame features to generate auxiliary temporal cues and then transform them back onto the network stem. We organize comprehensive ablation studies on the instantiation of ATMs, \textit{w.r.t} the choice of operations, temporal context range, structures of feature extractor, and the attached locations. Subsequently, we discover the composite of \emph{Subtraction} and \emph{Multiplication} achieve the best performance among the candidates. 
Finally, we evaluate the proposed method by instantiating it on both representative CNN and Transformer backbones (\ie, ResNet~\cite{resnet}, ViT~\cite{ViT}), to show its generality and superiority.
Our contributions are as follows:
\begin{itemize}
\item We explore the ascendency of employing fundamental arithmetic operations for temporal modeling, and propose an ad-hoc block named \textbf{A}rithmetic \textbf{T}emporal \textbf{M}odule (ATM) that integrates pair-wise temporal interactions between pairs of frame features.
\item We conduct a comprehensive exploration of the instantiation of ATMs on both CNN backbones and Vision Transformer backbones. 
\item We demonstrate that existing backbones plugged with ATM can achieve superior performance on a broad range of popular benchmarks. Specifically, our method achieves Top-1 accuracy of 65.6\% on Something-Something V1, 74.6\% on Something-Something V2, and 89.4\% on Kinetics-400, respectively.
\end{itemize}
\section{Related Work}
\label{sec:related}
\noindent\textbf{CNNs for Video Recognition.}
Convolutional neural networks (CNNs) have made significant contributions to the field of video recognition. Initially, temporal information was constructed using low-level streams (\eg, optical flow, motion vector, RGB difference)~\cite{two-stream-stresnet,two-stream,tsn,mv-cnn}, or through the use of segregated time sequences with recurrent networks that extracted features from an image backbone~\cite{LRCN,li2017temporal,yue2015beyond}.
As computational capacity rapidly improved, 3D CNNs~\cite{i3d,c3d} and factorized 2D+1D CNNs~\cite{slowfast,p3d,r2+1d,s3d} emerged as natural extensions of their 2D counterparts, allowing for the direct handling of 3D volumetric video data.
To further alleviate the computational cost of video classification networks, many studies~\cite{kwon2020motionsqueeze,li2020tea,tsm,teinet,liu2021tam,GST,GSM,tdn,wu2020MVFNet} have aimed to design efficient temporal modules and incorporate them into 2D CNNs~\cite{resnet,bn}.
There are also studies have developed efficient video architectures that do not rely on existing 2D CNNs, resulting in even more lightweight structures~\cite{feichtenhofer2020x3d,CSN,corrnet}.
Additionally, a few works \cite{dsanet,zhang2020v4d} have introduced 4D convolutional techniques to capture relationships between clips.
Another research direction is orthogonal to the aforementioned CNN-based video architecture, which focuses on dynamic inference~\cite{wu2019multi,wu2020dynamic,wu2019adaframe,tsqnet,nsnet,zhang2022twinnet} to achieve efficient recognition.

Among the aforementioned studies, \cite{kwon2020motionsqueeze,selfy,corrnet} employ correlation to compute the similarity between consecutive frames and incorporate it into the design of CNN-based temporal modeling. 
We consider these works to be relevant to ours since we view correlation as a specific instance of multiplication. 
However, apart from differences in implementation details, our goal is to explore the potential of various cues generated from the simple arithmetic operations and propose a universal temporal module applicable to both CNNs and Vision Transformers, rather than designing customized CNN-based video architectures.


\begin{figure*}[t]
\centering
\includegraphics[width=0.98\linewidth]{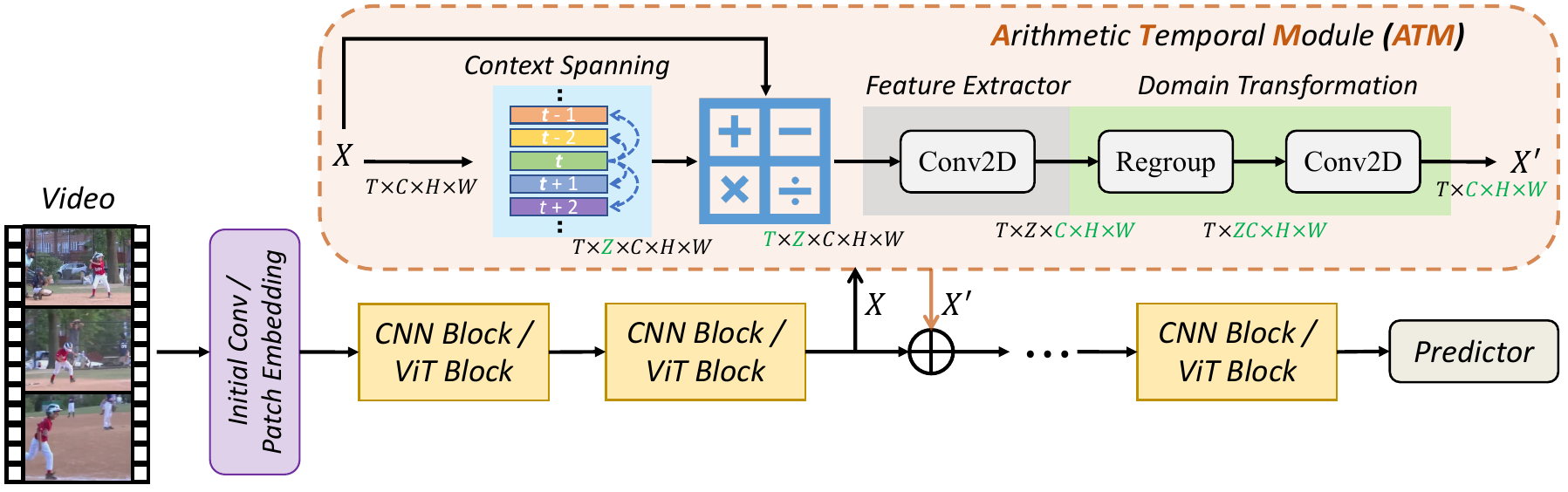}
\caption{The overall framework of the simple \textbf{A}rithmetic \textbf{T}emporal \textbf{M}odule (ATM) for temporal modeling. The key motivation behind ATM is to explore the potential of simple arithmetic operations to capture auxiliary temporal clues that may be embedded in current video features, without relying on the elaborate design. The ATM can be integrated into both vanilla CNN backbone (\eg, ResNet~\cite{resnet}) and Vision Transformer (\eg, ViT~\cite{ViT}) for video action recognition.
}
\label{fig:approach}
\end{figure*}

\noindent\textbf{Transformers for Video Recognition.}
Transformers~\cite{vaswani2017attention}, originally proposed for the language domain, have been adapted for vision tasks and have gained popularity in the computer vision community. 
Image Transformers \cite{ViT,DeiT,liu2021swin} have achieved great success, leading to the investigation of transformers for video recognition. 
Several works \cite{arnab2021vivit,timesformer,mvit,videoswin,VTN} propose different spatial and temporal attention designs using the large-scale ImageNet-21K~\cite{ridnik2021imagenet21k} pre-trained model or designing new models from scratch, demonstrating improved performance for spatiotemporal modeling for video recognition.



In addition to the evolution of video transformer architectures, large-scale pre-training has significantly boosted their popularity. Several studies \cite{arnab2021vivit,ryoo2021tokenlearner,MTV,cover} have achieved remarkable performance in video recognition task using the JFT-300M~\cite{JFT300M} or even JFT-3B~\cite{JFT3B} pre-trained image transformer. 
More recently, large-scale image-text pre-training has attracted wider attention. 
One representative work is CLIP~\cite{CLIP}, which leverages 400M image-text pairs (namely, WIT-400M) to train vision and text encoders using contrastive learning. This breakthrough has inspired several research endeavors that directly explore the application of CLIP models in text-video retrieval~\cite{fang2023uatvr,luo2022clip4clip,wu2022cap4video,zhao2022centerclip}, while others have also attempted to utilize CLIP for video recognition.
Based on whether CLIP's text encoder is utilized, these studies can be categorized into two lines: the video-text paradigm~\cite{ju2021prompting,x-clip,wang2021actionclip,text4vis,wu2023transferring,bike} and the video-only paradigm~\cite{evl,st-adapter,yang2023aim}. In our research, we only use CLIP's vision encoder without involving any textual knowledge for additional gains. 
Various CLIP variants have subsequently emerged. For example, Florence~\cite{yuan2021florence} utilizes 900M image-text pairs, while EVA-CLIP~\cite{sun2023eva} surpasses that by leveraging an impressive 2B image-text pairs (namely, Merged-2B) for pre-training.

\section{Approach}
In this section, we begin by providing an overview of the \textbf{A}rithmetic \textbf{T}emporal \textbf{M}odule (ATM), followed by the introduction of its structure. We then provide a detailed explanation of the arithmetic operations employed in ATM.

\subsection{Overview of ATM}
The key motivation behind ATM is to leverage the potential of simple arithmetic operations that do not require parameters to capture additional temporal information present in current frame features. This approach provides a more straightforward and interpretable way to capture additional temporal clues without the need for elaborate design. By integrating ATM into various backbones, including the popular CNN - ResNet~\cite{resnet} and Vision Transformer - ViT~\cite{ViT}, effective temporal modeling can be achieved.

Specifically, ATM consists of several stages. Firstly, context spanning is performed on temporal-irrespective frame features to construct context frames. Subsequently, arithmetic operations (\faPlus, \faMinus, \faTimes, \faDivide) are applied to the features of each frame and its context frames to obtain temporal clues. These temporal clues are then fed into a feature extractor to extract their features. Finally, the temporal clues are projected back to the original temporal-irrespective domain through domain transformation, and are connected to the original features in a residual form. 
The overall framework is illustrated in \Cref{fig:approach}.

\subsection{Arithmetic Temporal Module (ATM)}\label{sec:approach:atm}
\paragraph{The Structure of ATM.}
Here we describe the specific structure of the ATM. Given a video, we sample $T$ frames for video recognition. To capture the temporal information, we propose to explicitly utilize the relationship between context frames to capture the temporal information. Specifically, we use \textbf{\emph{Context Spanning}} to construct $Z$ context frames for each frame, where $Z$ defines the context range. For clarity, we define the pair-wise \textbf{\emph{Arithmetic Operation}} (any of \faPlus, \faMinus, \faTimes, \faDivide) as a function $\psi(\cdot, \cdot)$ that performs point-to-point operations on two frame features in the spatial dimensions $H$ and $W$. The output of the arithmetic operation is formulated as follows:
\begin{equation}
\begin{split}
    X^{\texttt{OUT}} = \texttt{Concatenate} & \Big( \big[ \psi (X_{t}, X_{z}), \ldots \big] \Big), \\
    & t = 1, 2 \ldots T, z \in \mathcal{Z} 
\end{split}
\end{equation}
where $X_t \in \mathbb{R}^{C\times H\times W}$ represents the feature of the anchor frame, while $X_z \in \mathbb{R}^{C\times H\times W}$ represents the feature of one of its spanned context frame. $C, H$, and $W$ denote the number of channels, height, and width of a channel slice, respectively. $\mathcal{Z}$ denotes the index set of context frames. If we define $Z$ as the number of context frames in $\mathcal{Z}$, the output feature $X^{\texttt{OUT}}$ becomes a tensor of size $\mathbb{R}^{T\times Z\times C\times H\times W}$. The context frames are chosen based on context range $Z$ to interact with the anchor frame:
\begin{itemize}
    \item When $Z=1$, the context is chosen as the next frame, $\mathcal{Z} = \{ t+1 \}$.
    \item When $Z=2,4,6 \ldots$, the context is chosen as the previous $\frac{Z}{2}$ and next $\frac{Z}{2}$ frames, \eg, $\mathcal{Z} = \{ t-1, t+1 \}$ for $Z=2$.
\end{itemize}

After performing the arithmetic operation for temporal interaction, we obtain a composite temporal signal of size $\mathbb{R}^{T\times Z\times C\times H\times W}$, where the $T \times Z$ matrix records the pair-wise temporal cues for each frame. We then employ a \textbf{\emph{Feature Extractor}} that extracts spatially dependent representations while preserving the $T\times Z$ matrix. In practice, we implement the feature extractor using a stack of two $3 \times 3$ convolutions over the spatial axes.

Finally, we perform \textbf{\emph{Domain Transformation}} to project the representations back to the temporal-irrespective domain. Specifically, we regroup the pair-wise interaction axis $Z$ with the spatial channel $C$. To reduce the dimensionality of the representations from $ZC$ to $C$, we employ a $1 \times 1$ convolution over the channel dimension. The resulting features are then connected to the original features $X \in \mathbb{R}^{T\times C\times H\times W}$ in a residual form, providing effective temporal information to the original features. 

\vspace{-4mm}
\paragraph{Integrating ATM into CNNs and Vision Transformers.}
To showcase the versatility of our simple ATM, we integrate it into existing image backbones, enabling powerful temporal modeling. We specifically select two widely-used backbone architectures: CNNs and Vision Transformers. In our experiments, we incorporate ATM into vanilla ResNet~\cite{resnet} and ViT~\cite{ViT}, which are representative architectures in their respective domains. For ResNet, we insert one ATM after the last residual block of res$_3$. For ViT, we integrate one ATM after the $7$-th layer of ViT-B (which has 12 layers) and after the $14$-th layer of ViT-L (which has 24 layers).


\subsection{Arithmetic Operations}\label{sec:approach:wwhnet}
In \Cref{sec:approach:atm}, we present the structure of our ATM. Now, let's delve into the instantiation of the arithmetic operations employed within ATM. Specifically, we consider four fundamental arithmetic operations: \faPlus, \faMinus, \faTimes, \faDivide.

(1) \emph{Addition} (\faPlus): We use the addition operation to compute the accumulated signals of actions. Intuitively, it captures the spatial occupancy of feature representations, \ie, patterns of a series of actions, which is helpful in understanding actions like gestures. The addition of $A$ with $B$ at spatial location $(c, h, w)$ is computed as
\begin{equation}
{Addition}(A,B)_{(c, h, w)} = A(c, h, w) + B(c, h, w),
\end{equation}
yielding an \emph{Addition} tensor of size $C$\x$H$\x$W$.

(2) \emph{Subtraction} (\faMinus): The subtraction operation is widely used to estimate optical flow, where it represents the magnitude of the spatial feature change. Here, we use the subtraction operation to infer the smoothness of the change by concatenating the subtraction, which helps us understand the action dynamics. The subtraction of $A$ with $B$ at spatial location $(c, h, w)$ is computed as
\begin{equation}
{Subtraction}(A,B)_{(c, h, w)} = A(c, h, w) - B(c, h, w),
\end{equation}
yielding a \emph{Subtraction} tensor of size $C$\x$H$\x$W$.

(3) \emph{Multiplication} (\faTimes): The spatial multiplication between two features can be considered as the apparent similarity of spatial features. The similarity signals capture the likelihood of the spatial appearance of a feature in the interval of the paired frames, and can be useful in understanding the dynamics of actions. In this paper, we utilize the local multiplication of a $P \times P$ neighborhood instead of global multiplication. This operation is also referred to as correlation in some cases. Concretely, we compute the multiplication of $A$ with $B$ at spatial location $(h, w)$ as
\begin{equation}
\begin{split}
{Multiplication}(A,B)_{(h,w)} = & A(h,w)   \cdot B(h+i,w+j), \\ 
& (i,j)\in {P\times P}.
\end{split}
\end{equation}
Here, we use ${P\times P}$ to denote the relative spatial location set of neighborhoods. For $P = 2k+1$, $\{P\times P\} \equiv \{-k, \ldots, k\} \times  \{-k, \ldots, k\}$, where $k$ represents the maximum offset within the neighborhood. We set $P$ to 9 in this paper empirically. Thus, we obtain a \emph{Multiplication} tensor of size $P^2$\x$H$\x$W$, which is transformed to the size of the original feature $C$\x$H$\x$W$ via several convolutional layers.

(4) \emph{Division} (\faDivide): The element-wise division operation amplifies the changes in features between a frame and its anchor frame. This operation functions similarly to subtraction, but the magnitude of the difference is standardized, allowing even features with small numerical values to result in a large quotient. To stabilize the training process, we employ the subtraction between the \texttt{log} value of the features, which can be expressed as
\begin{equation}
\begin{split}
{Division}(A,B)_{(h,w)} = & log (A(c, h, w) + \epsilon) - \\
 & log (B(c, h, w) + \epsilon), 
\end{split}
\end{equation}
where $\epsilon$ is a stability parameter set to $1$ in our experiments, which avoids a non-positive value in the logarithm operator.

\section{Experiments}

\subsection{Datasets and Evaluation Metrics}
We evaluate our ATM on five video recognition benchmarks, including Something-Something (SS) V1\&V2~\cite{sth-sth}, Kinetics-400 (K400)~\cite{kay2017kinetics}, ActivityNet~\cite{caba2015activitynet}, and Charades~\cite{charades}. The SS V1 contains 110k videos, while V2 contains 220k video clips across 174 fine-grained classes. In contrast, Kinetics-400 is another widely-used large-scale video dataset, consisting of 300k video clips across 400 human action categories.
In \emph{Appendix}, we also demonstrate the generalization ability of our ATM by evaluating it on smaller datasets, \eg, ActivityNet and Charades. 

Our results are reported according to the official evaluation metrics. We report top-1 and top-5 accuracy for K400 and SS V1\&V2, respectively. For ActivityNet and Charades, we report the mean average precision (mAP). In addition, we report the computational cost (in FLOPs) to provide a comprehensive understanding of model complexity. Note that we only employ RGB frames for all datasets.

\subsection{Implementation Details}
\noindent\emph{\textbf{Training.}} In this paper, we employ ResNet~\cite{resnet} and ViT~\cite{ViT} as representative backbones for CNNs and Vision Transformers, respectively. For CNN-based instantiation, we utilize the widely-used ImageNet-1K~\cite{deng2009imagenet} pre-trained ResNet, which aligns with common practices in prior studies~\cite{tsm,wu2020MVFNet,tdn}. In the case of Transformer-based instantiation, we follow the recent trend~\cite{evl,st-adapter,yang2023aim} by adopting the visual encoder of CLIP~\cite{CLIP} as the backbone.
Our models are trained using either 8 or 16 frames, and we provide detailed training configurations in \emph{Appendix}.

%

\noindent\emph{\textbf{Inference.}}
To trade off accuracy and speed, we consider two evaluation protocols. 
(1) \emph{Single View}: We use only 1 clip per video and the center crop for efficient evaluation.
(2) \emph{Multiple Views}: It is a common practice~\cite{i3d,slowfast,wu2020MVFNet} to sample multiple clips per video with several spatial crops to get higher accuracy. 
For readers' reference, we present our results with various views in the \emph{Appendix}.

\begin{table*}[t]
\centering
    \begin{subtable}[t]{0.25\textwidth}
    \centering
      \begin{tabular}[t]{ccccc}
      \toprule
      \faPlus & \faMinus & \faTimes & \faDivide & \textbf{Top-1}  \\
      \midrule
       - & - & - & - & 16.5\% \\ \hdashline
       \cmark & - & - & - & 41.1\% \\
        - & \cmark & - & - & 43.6\% \\
        - & - & \cmark & - & \base{\textbf{46.5\%}}  \\
        - & - & - & \cmark & 43.1\%  \\
      \bottomrule
      \end{tabular}
      \caption{Study on the effects of four arithmetic operations independently.}
      \label{table:op}
    \end{subtable}
    \hspace{2mm}
    \begin{subtable}[t]{0.25\textwidth}
        \centering
          \begin{tabular}[t]{ccccc}
          \toprule
      \faPlus & \faMinus & \faTimes & \faDivide & \textbf{Top-1} \\
      \midrule
      \cmark & - & \cmark & - & 47.8\% \\
      - & \cmark & \cmark & - & \base{\textbf{48.2\%}} \\
      - & - & \cmark & \cmark & 47.3\% \\
      \cmark & \cmark & - & - & 44.6\%  \\ 
      \cmark & \cmark & \cmark & - & 48.1\%  \\ 
          \bottomrule
          \end{tabular}
          \caption{Study on the complementarity among four temporal interactions.}
        \label{tab:fuse}
    \end{subtable}
    \hspace{2mm}
    \begin{subtable}[t]{0.2\textwidth}
        \centering
        \begin{tabular}[t]{ccc}
         \toprule
         \textbf{$Z$} & \textbf{Top-1} & \textbf{FLOPs} \\
         \midrule
         \demph{0} & \demph{16.5\%} & \demph{14.5G} \\ \hdashline
         $1$ & 43.7\% & 15.4G \\
         $2$ & 47.0\% & 16.5G \\
         $4$ & \base{\textbf{48.2\%}} & 17.7G \\
         $6$ & 48.0\% & 18.8G\\
         \bottomrule
         \end{tabular}
         \caption{Study on the temporal context range $Z$.}
        \label{tab:context_range}
    \end{subtable}
    \hspace{2mm}
    \begin{subtable}[t]{0.22\textwidth}
        \centering
        \begin{tabular}[t]{lc}
         \toprule
         \textbf{Method} & \textbf{Top-1} \\
         \midrule
         T-Conv & 44.3\%  \\
         ATM (\faMinus, \faTimes) & 48.2\% \\
         ATM + T-Conv & \base{\textbf{48.7\%}}  \\
         \bottomrule
         \end{tabular}
         \caption{Exploring compatibility of ATM with Temporal Convolution (T-Conv).}
        \label{tab:tconv}
    \end{subtable}    
    \\[7pt]
    \begin{subtable}[t]{0.22\textwidth}
        \centering
      \begin{tabular}[t]{lcc}
      \toprule
       & \textbf{Top-1}  \\
      \midrule
       FC & 47.8\% \\
       MLP  & 48.1\%  \\ 
       Conv Stack & \base{\textbf{48.7\%}}  \\
      \bottomrule
      \end{tabular}
    \caption{Study on various feature extractor in ATM. MLP: FC\x3. Conv Stack: Conv2D\x2.
    }
      \label{table:feature_extractor}
    \end{subtable}   
    \hspace{3mm}
    \begin{subtable}[t]{0.25\textwidth}
        \centering
        \begin{tabular}[t]{lcc}
        \toprule
         \textbf{Stage}  & \textbf{Top-1} & \textbf{FLOPs} \\
         \midrule
         res$_2$  & 48.8\% & 26.5G \\
         res$_3$  & \base{\textbf{48.7\%}} & 17.7G \\
         res$_4$  & 47.6\% & 17.9G \\
         res$_5$  & 43.1\% & 15.5G \\ 
         \bottomrule
         \end{tabular}
         \caption{Study on the location of ATM.}
        \label{tab:position}
    \end{subtable}
    \hspace{3mm}
    \begin{subtable}[t]{0.46\textwidth}
        \centering
      \begin{tabular}[t]{ccccc}
      \toprule
      \textbf{Model} & \textbf{\#Frame} & \textbf{Backbone} & \textbf{Top-1} & \textbf{Top-5} \\
      \midrule
      \demph{TSM~\cite{tsm}} & \demph{8} & \demph{ResNet-50} & \demph{47.1\%} & \demph{76.2\%} \\ \midrule
      \multirow{3}*{Ours} & 8 & ResNet-18 & 48.7\% & 77.7\% \\
       & 8 & ResNet-50 & 53.9\% & 81.8\% \\
       & 16 & ResNet-50 & 56.3\% & 83.4\% \\
      \bottomrule
      \end{tabular}
      \caption{Study on deeper ResNet backbones and more frames.}
      \label{table:backbone}
    \end{subtable}  
    \vspace{-2mm}
    \caption{Ablation studies with the representative \textbf{CNN backbone - ResNet}~\cite{resnet} on Something-Something V1. Unless otherwise specified, all models use ResNet18 with 8 frames under the single view protocol.}
    \label{tab:ab_study_resnet}
\end{table*}

\begin{table*}[htbp]
\centering
    \begin{subtable}[t]{0.26\textwidth}
    \centering
      \begin{tabular}[t]{ccccc}
      \toprule
      \faPlus & \faMinus & \faTimes & \faDivide & \textbf{Top-1}  \\
      \midrule
       - & - & - & - & 27.3\% \\ \hdashline
       \cmark & - & - & - & 51.3\% \\
        - & \cmark & - & - & 53.8\% \\
        - & - & \cmark & - & \base{\textbf{54.5\%}}  \\
        - & - & - & \cmark & 53.7\%  \\
      \bottomrule
      \end{tabular}
      \caption{Study on the effects of four arithmetic operations independently.}
      \label{table:op_vit}
    \end{subtable}
    \hspace{2mm}
    \begin{subtable}[t]{0.2\textwidth}
        \centering
        \setlength{\tabcolsep}{2.0pt}
          \begin{tabular}[t]{lc}
          \toprule
         \textbf{Method} & \textbf{Top-1} \\ \midrule 
         ATM (\faMinus, \faTimes) & 56.1\% \\
         T-Conv & 51.3\% \\ \midrule
         ATM + T-Conv & \base{\textbf{58.1\%}} \\
          \bottomrule
          \end{tabular}
          \caption{Exploring compatibility of ATM with Temporal Convolution (T-Conv).}
        \label{tab:fuse_vit}
    \end{subtable}
    \hspace{2mm}
    \begin{subtable}[t]{0.19\textwidth}
        \centering
        \begin{tabular}[t]{lc}
        \toprule
         \textbf{Position}  & \textbf{Top-1}  \\
         \midrule
         layer\{2\}  & 36.9\%  \\
         layer\{4\}  & 55.7\% \\
         layer\{6\}  & 57.5\%  \\
         layer\{7\}  & \base{\textbf{58.1\%}} \\ 
         layer\{8\}  & 57.1\%  \\
         layer\{10\}  & 55.7\%  \\
         \bottomrule
         \end{tabular}
         \caption{The location of ATM.}
        \label{tab:vit_position}
    \end{subtable}    
    \hspace{2mm}
    \begin{subtable}[t]{0.28\textwidth}
        \centering
      \begin{tabular}[t]{ccc}
      \toprule
       \textbf{\#Frame} & \textbf{Backbone} & \textbf{Top-1} \\
      \midrule
       8 &  ViT-B & 58.1\%  \\
       16 & ViT-B & 59.5\% \\ 
       32 & ViT-B & 60.9\% \\  
        8 & ViT-L & 61.9\% \\
        16 & ViT-L & 63.3\% \\
      \bottomrule
      \end{tabular}
      \caption{Study on larger ViT backbones and more frames.}
      \label{table:backbone_vit}
    \end{subtable}  
    \vspace{-2mm}
    \caption{Ablation studies with representative \textbf{Vision Transformer - ViT}~\cite{ViT} on Something-Something V1. Unless otherwise specified, all models use ViT-B/16 with 8 frames under the single view protocol.}
    \label{tab:ab_study_vit}
\end{table*}


\subsection{Ablation Studies}\label{sec:exp:ablation}

We conduct ablation studies on both ResNet and ViT to evaluate the temporal modeling capability of our ATM. The Something-Something dataset involves complex object-object and human-object interactions, which require capturing strong temporal relationships and motion cues for accurate recognition. Therefore, recognition based solely on scene information is challenging for this dataset. The results of our studies are presented in Table~\ref{tab:ab_study_resnet} and Table~\ref{tab:ab_study_vit}.

\noindent\textbf{Four Arithmetic Operations.} 
We investigate the impact of four arithmetic operations - Addition (\faPlus), Subtraction (\faMinus), Multiplication (\faTimes), and Division (\faDivide) - on the performance of our proposed method for temporal modeling. We also compare the results of each operation with a baseline approach that uses mean pooling to model the temporal information of videos, focusing only on appearance features.
The results shown in Tables~\ref{table:op} and ~\ref{table:op_vit} demonstrate that each arithmetic operation significantly outperforms the baseline, indicating that simple arithmetic operations can provide strong temporal cues for effective temporal modeling. We also observe that \faTimes, \faMinus, and \faDivide~lead to relatively better results compared to \faPlus. These results are consistent across both ResNet and ViT backbones.

\noindent\textbf{Fusion of Multiple Temporal Cues.}
We investigate the redundancy and complementarity of various sources of temporal information by stacking different instances of ATM.
Since \faTimes~achieves the highest performance, we primarily adopt it and combine it with other operations. Based on Tables \ref{table:op} and \ref{tab:fuse}, we observe that all three other operations complement \faTimes~to a certain extent, with \faMinus~and \faTimes~showing the most prominent complementary effect. We also find that \faPlus~and \faMinus~can complement each other. Therefore, we combine \faPlus, \faMinus, and \faTimes, but do not observe any additional improvement.
Similarly, as shown in Tables~\ref{table:op_vit} and \ref{tab:fuse_vit}, the fusion of \faMinus~and \faTimes~on ViT improves their individual results.


\begin{table*}[t]
\centering
\scalebox{0.92}{
\begin{tabular}{lccccccc}
\toprule
\multirow{2}*{\textbf{Methods}} & \multirow{2}*{\textbf{Year}} & \multirow{2}*{\textbf{Backbones}} & \multirow{2}*{\textbf{Pre-training}} &\multirow{2}*{\textbf{Frames\x Crops\x Clips}} & \multirow{2}*{\textbf{GFLOPs}} & \multirow{2}*{\textbf{SSV1}} & \multirow{2}*{\textbf{SSV2}} \\
 &  &  &  & & & \\
\midrule
TSM~\cite{tsm} & ICCV'2019 & ResNet50 & ImageNet-1K & 16\x1\x1 & 65 & 47.2\% & 63.4\% \\
TSM$_{En}$~\cite{tsm} & ICCV'2019 & ResNet50 & ImageNet-1K  & (8+16)\x1\x1 & 98 & 49.7\% & N/A\\
TEINet$_{En}$~\cite{teinet} & AAAI'2020 & ResNet50 & ImageNet-1K  & (8+16)\x1\x1 & 99 & 52.5\% & 66.5\% \\
TEA~\cite{li2020tea} & CVPR'2020 & ResNet50 & ImageNet-1K  & 16\x3\x10 & 70\x30 & 52.3\% & N/A\\
MSNet~\cite{kwon2020motionsqueeze} & ECCV'2020  & ResNet50 & ImageNet-1K  & 16\x1\x1 & 67 & 52.1\% & 64.7\% \\
MVFNet$_{En}$~\cite{wu2020MVFNet} & AAAI'2021 & ResNet50 & ImageNet-1K  & (8+16)\x3\x2 & 99\x6 & 54.0\% & 66.3\% \\
TDN$_{En}$~\cite{tdn} & CVPR'2021 & ResNet50  & ImageNet-1K & (8+16)\x1\x1 & 108 & 55.1\% & 67.0\%\\
SELFY$_{En}$~\cite{selfy} & ICCV'2021 & ResNet50  & ImageNet-1K & (8+16)\x1\x1 & 114 & 55.8\% & 67.4\%\\
\midrule
 Ours & \multirow{3}*{ICCV'2023} & ResNet50 & ImageNet-1K & 16\x1\x1 & 74 & 56.3\% & 67.4\%\\
 Ours & & ResNet50 & ImageNet-1K & 32\x1\x1 & 148 & 57.1\% & 68.4\%\\ 
 Ours$_{En}$ & & ResNet50 & ImageNet-1K & (8+16)\x1\x1 & 111 & 57.6\% & 68.3\% \\
\midrule\midrule
CorrNet~\cite{corrnet} & CVPR'2020 & ResNet101 & ImageNet-1K & 32\x3\x10 & 224\x30 & 51.7\% & N/A\\
 TDN~\cite{tdn} & CVPR'2021 & ResNet101 & ImageNet-1K & 16\x1\x1 & 132 & 55.3\% & 66.9\%\\ 
 TDN$_{En}$~\cite{tdn} & CVPR'2021 & ResNet101 & ImageNet-1K & (8+16)\x1\x1 & 198 & 56.8\% & 68.2\%\\ 
 \midrule
 Ours & \multirow{3}*{ICCV'2023} & ResNet101 & ImageNet-1K & 16\x1\x1 & 134 & 57.2\% & 68.2\%\\
 Ours$_{En}$ & & ResNet101  & ImageNet-1K & (8+16)\x1\x1 & 201 & 58.6\% & 69.4\% \\ 
 Ours$_{En}$ & & ResNet101 & ImageNet-1K  & (8+16+32)\x1\x1 & 469 & \textbf{60.0\%} & \textbf{70.8\%} \\ 
 \midrule\midrule
 ViViT~\cite{arnab2021vivit} & ICCV'2021 & L/16\x2 FE & IN-21K & 32\x3\x14 & 47600 & N/A & 65.4\% \\
MTV~\cite{MTV} & CVPR'2022 & MTV-B (320$\uparrow$) & IN-21K & 32\x3\x4 & 11200 & N/A & 68.5\% \\
 CoVeR~\cite{cover} & ArXiv'2022 & TSFormer (448$\uparrow$) & JFT-3B & 16\x3\x1 & 17600 & N/A & 70.8\% \\
 EVL~\cite{evl} & ECCV'2022 & ViT-B/16 & WIT-400M & 32\x3\x1 & 2047 & N/A & 62.4\% \\
 EVL~\cite{evl} & ECCV'2022 & ViT-L/14 & WIT-400M & 32\x3\x1 & 9641 & N/A & 66.7\% \\
 AIM~\cite{yang2023aim} & ICLR'2023 & ViT-B/16 & WIT-400M & 32\x3\x1 & 2496 & N/A & 69.1\% \\
 AIM~\cite{yang2023aim} & ICLR'2023 & ViT-L/14 & WIT-400M & 32\x3\x1 & 11508 & N/A & 70.6\% \\
 ST-Adapter~\cite{st-adapter} & NeurIPS'2022 & ViT-L/14 & WIT-400M & 32\x3\x1 & 8248 & N/A & 72.3\% \\ 
 UniFormerV2~\cite{li2022uniformerv2} & ICCV'2023 & ViT-L/14 & WIT-400M & 32\x3\x1 & 5200 & 62.7\% & 73.0\% \\ 
 \midrule
  Ours & \multirow{4}*{ICCV'2023} & ViT-B/16 & WIT-400M & 32\x3\x1 & 378\x3 & 61.3\% & 71.9\%\\
  Ours & & ViT-L/14 & WIT-400M & 16\x3\x1 & 842\x3 & 63.7\% & 73.2\%\\
  Ours & & ViT-L/14 & WIT-400M & 16\x3\x2 & 842\x6 & 64.0\% & 73.5\%\\
  Ours & & ViT-L/14 & Merged-2B & 16\x3\x2 & 842\x6 & \textbf{65.6\%} & \textbf{74.6\%}\\
 \bottomrule
\end{tabular}
}
\vspace{-1mm}
\caption{Comparison with the SOTAs on Something-Something (SS) V1 \& V2. Unless otherwise specified, all models use a spatial size of 224\x224 as the input. Top-1 accuracy is reported. $En$ denotes the ensemble of multiple models with different numbers of input frames.}
\label{tab:sth_sota}
\end{table*}

\noindent\textbf{Temporal Context Range.}
We investigate the impact of varying the temporal context range $Z$ on pair-wise operations and present the results in Table~\ref{tab:context_range}. The context range determines the scope of obtaining temporal cues for each frame from its adjacent frames. For instance, setting $Z=4$ implies that each frame undergoes arithmetic operations with its two preceding and two succeeding frames. We observe that a larger context range can enhance the non-context model, resulting in noticeable performance gain. However, increasing $Z$ beyond 4 did not yield a significant improvement. Hence, we set $Z$ to 4 as the default value.

\noindent\textbf{Complementarity with Temporal Convolution.}
Temporal convolution (T-Conv) is a common and efficient temporal operation used in previous methods~\cite{li2020tea,tsm,teinet,wu2020MVFNet}. It performs a 3\x1\x1 depth-wise convolution along the temporal dimension. To incorporate T-Conv into our models, we insert T1D into each block of ResNet's res4 and res5, and apply T1D to the [CLS] token in all layers of ViT, which is a commonly used practice.
We investigate the complementarity between ATM and T-Conv. Table~\ref{tab:tconv} and \ref{tab:fuse_vit} show that although T-Conv performs worse than ATM on both ResNet and ViT, combining ATM with T-Conv further improves performance. Therefore, we include the lightweight T-Conv in our default setting for subsequent experiments.

\noindent\textbf{Feature Extractor.}
We also investigate the impact of different feature extractors in our framework. We experiment with three common instantiations: Fully Connected layer (FC), MultiLayer Perceptron (MLP), and a stack of Convolutional layers (Conv Stack). As shown in Table \ref{table:feature_extractor}, the Conv Stack achieves the highest accuracy of 48.7\%, outperforming the other two options with 47.8\% and 48.1\% for FC and MLP, respectively. Therefore, we select the Conv Stack as the default feature extractor in our experiments.

\noindent\textbf{The Optimal Position for ATM.}
We investigate the optimal position for inserting the ATM in ResNet and ViT.
We denote the conv2\_x to conv5\_x layers of ResNet architecture as res$_2$ to res$_5$. To gradually incorporate the ATM into ResNet-18, we add it from res$_2$ to res$_5$. For example, res$_2$ denotes that one ATM is inserted after the last res block of res$_2$. 
As shown in Table~\ref{tab:position}, adding ATM to res${_2}$ or res${_3}$ results in significant improvements. Performance is minimally impacted when ATM is added to res${_2}$ or res${_3}$, and thus we use ATM in res${_3}$ for efficiency. Regarding ViT, we also observe that inserting ATM into the middle layers leads to better performance, as shown in Table~\ref{tab:vit_position}. Thus, we insert one ATM after the $7$-th layer of ViT-B (which has 12 layers) and after the $14$-th layer of ViT-L (which has 24 layers).

\noindent\textbf{More Instantiations.}
We evaluate our method using various visual encoders and more input frames. Tables~\ref{table:backbone} and \ref{table:backbone_vit} present the results of different instantiations based on ResNet and ViT, respectively. Notably, our method with ResNet-18 outperforms TSM-ResNet50~\cite{tsm} (48.7\% \emph{vs.} 47.1\%). In general, we observe that deeper backbones can achieve better performance, and increasing the number of frames can lead to additional improvements.




\begin{table*}[t]
\centering
\scalebox{0.92}{
\begin{tabular}{lccccccc}
\toprule
\textbf{Methods} & \textbf{Year} &  \textbf{Backbones} & \textbf{Pre-training} & \textbf{Frames\x Crops\x Clips} & \textbf{GFLOPs} & \textbf{Top-1}  & \textbf{Top-5} \\
\midrule
TSM~\cite{tsm} & ICCV'2019 & ResNet-50 & ImageNet-1K & 8\x3\x10 & 33\x30 & 74.1\% & 91.2\%\\
TEINet~\cite{teinet} & AAAI'2020 & ResNet-50& ImageNet-1K & 8\x3\x10 & 33\x30 & 74.9\% & 91.8\% \\ 
TEA~\cite{li2020tea} & CVPR'2020 & ResNet-50& ImageNet-1K & 8\x3\x10 & 33\x30 & 75.0\% & 91.8\% \\
MVFNet~\cite{wu2020MVFNet} & AAAI'2021 & ResNet-50 & ImageNet-1K & 8\x3\x10  & 32.9\x30 & 76.0\% & 92.4\% \\
TANet~\cite{liu2021tam} & ICCV'2021 & ResNet-50 & ImageNet-1K & 8\x3\x10 & 43\x30 & 76.3\% & 92.6\%\\
TDN~\cite{tdn} & CVPR'2021 & ResNet-50 & ImageNet-1K & 8\x3\x10 & 36\x30 & 76.6\%  & 92.8\% \\
\midrule
Ours & \multirow{2}*{ICCV'2023} & ResNet-50 & ImageNet-1K & 8\x3\x10 & 37\x30 & 77.0\%  & 92.9\% \\
Ours & & ResNet-50 & ImageNet-1K & 16\x3\x10 & 74\x30 & \textbf{77.6\%}  &  \textbf{93.2\%} \\
\midrule\midrule
NL+I3D~\cite{nonlocal} & CVPR'2018 & ResNet-101 & ImageNet-1K & 128\x3\x10 & 359\x30 & 77.7\% & 93.3\% \\ 
SmallBig~\cite{li2020smallbignet} & CVPR'2020 & ResNet-101 & ImageNet-1K & 32\x3\x4 & 418\x12 & 77.4\% & 93.3\% \\
MVFNet~\cite{wu2020MVFNet} & AAAI'2021 & ResNet-101 & ImageNet-1K & 16\x3\x10 & 125\x30 & 78.4\% & 93.4\% \\ 
TDN~\cite{tdn} & CVPR'2021 & ResNet-101 & ImageNet-1K & 16\x3\x10  & 132\x30 & 78.5\% & 93.9\% \\ 
\midrule
Ours & \multirow{3}*{ICCV'2023} & ResNet-101 & ImageNet-1K & 16\x3\x10 & 134\x30 & 78.8\%  &  93.7\% \\
Ours & & ResNet-152 & ImageNet-1K & 16\x3\x10 & 191\x30 & 79.4\%  &  93.7\% \\
Ours$_{En}$ &   & R101+R152 & ImageNet-1K & (16+16)\x3\x10  & 326\x30 & \textbf{80.5\%} & \textbf{94.4\%} \\ 
\midrule\midrule
ViViT~\cite{arnab2021vivit} & ICCV'2021 & H/16\x2 & JFT-300M & 32\x4\x3 & 8316\x12 & 84.8\% & 95.8\% \\
MTV~\cite{MTV} & CVPR'2022 & MTV-H & JFT-300M & 32\x4\x3 & 3706\x12 & 85.8\% & 96.6\% \\
Florence~\cite{yuan2021florence} & ArXiv'2021 & Florence (384$\uparrow$) & FLD-900M & 32\x4\x3 & N/A & 86.5\% & 97.3\% \\
ST-Adapter~\cite{st-adapter} & NeurIPS'2022  & ViT-L/14 & WIT-400M & 32\x1\x3 & 8248 & 87.2\% & 97.6\% \\
AIM~\cite{st-adapter} & ICLR'2023  & ViT-L/14 & WIT-400M & 32\x1\x3 & 11208 & 87.5\% & 97.7\% \\
EVL~\cite{evl} & ECCV'2022  & ViT-L/14 & WIT-400M & 32\x1\x3 & 8088 & 87.3\% & - \\
EVL~\cite{evl} & ECCV'2022  & ViT-L/14 (336$\uparrow$) & WIT-400M & 32\x1\x3 & 18196 & 87.7\% & - \\
X-CLIP~\cite{x-clip} & ECCV'2022 &  ViT-L/14 (336$\uparrow$) $^\dag$ & WIT-400M & 16\x4\x3 & 3086\x12 & 87.7\% & 97.4\% \\
Text4Vis~\cite{wu2023transferring} & IJCV'2023 &  ViT-L/14 (336$\uparrow$) $^\dag$  & WIT-400M & 32\x4\x3 & 3829\x12 & 88.4\% & 98.0\% \\
BIKE~\cite{bike} & CVPR'2023 &  ViT-L/14 (336$\uparrow$) $^\dag$  & WIT-400M & 32\x4\x3 & 3728\x12 & 88.6\% & 98.3\% \\
\midrule

Ours & \multirow{4}*{ICCV'2023} & ViT-L/14 & WIT-400M & 8\x3\x4 & 421\x12 & 87.3\%  &  97.4\% \\
Ours & & ViT-L/14 & WIT-400M & 32\x3\x4 & 1684\x12 & 88.0\%  &  97.6\% \\
Ours & & ViT-L/14 (336$\uparrow$) & WIT-400M & 32\x3\x4 & 3784\x12 & 88.2\%  &  97.9\% \\
Ours & & ViT-L/14 (336$\uparrow$) & Merged-2B & 32\x3\x4 & 3784\x12 & \textbf{89.4\%}  &  \textbf{98.3\%} \\

\bottomrule
\end{tabular}
}
\caption{Comparison with the SOTAs on Kinetics-400. Unless otherwise specified, all models use a spatial size of 224\x224 as the input. ``IN" indicates ImageNet. $^\dag$ denotes that the method utilizes textual knowledge from the text encoder of CLIP~\cite{CLIP}.
}
\label{tab:k400_sota}
\end{table*}

\subsection{Main Results}
Until 2021, ResNet was the primary backbone for video recognition methods. However, since 2022, the landscape has shifted with the advent of new techniques utilizing vision transformers that incorporate large-scale pre-training~\cite{CLIP,ridnik2021imagenet21k,JFT300M,sun2023eva,JFT3B}. In this section, we present the results of our method on both ResNet and ViT backbones, and compare them with previous works.


\noindent\textbf{Results on Something-Something V1 \& V2}.
We summarize our results in Table~\ref{tab:sth_sota}. 
To ensure fair comparisons with CNN-based methods, we evaluate our ATM using different configurations, including various backbones (ResNet-50/101) and numbers of input frames (8/16). Our ATM consistently outperforms previous state-of-the-art works presented at top-tier conferences on both datasets. For instance, on SS V1, we achieve better results than TDN~\cite{tdn} using both ResNet-50 (\textbf{57.6\%} \emph{vs.} 55.1\%) and ResNet-101 (\textbf{58.6\%} \emph{vs.} 56.8\%) backbones. Similar conclusions can be drawn for SS V2.
We also compare our method with recent works that utilize large-scale pre-trained video transformers~\cite{evl,yang2023aim,st-adapter}, which use the same CLIP pre-trained backbone. Our approach shows significant improvements in accuracy, with an absolute improvement of 9\% (\textbf{71.9\%} \emph{vs.} 62.4\%), while utilizing fewer FLOPs (1188G \emph{vs.} 2047G) compared to EVL~\cite{evl} with the same number of frames. In comparison to more recent methods such as AIM~\cite{yang2023aim}, ST-Adapter~\cite{st-adapter} and UniFormerV2~\cite{li2022uniformerv2}, our approach consistently outperforms them. 
Moreover, when we significantly increase the scale of the pre-training data from 400 million to 2 billion samples (namely Merged-2B~\cite{sun2023eva}, which merges 1.6 billion samples from the LAION-2B~\cite{schuhmann2022laion} dataset with 0.4 billion samples from the COYO-700M~\cite{kakaobrain2022coyo-700m} dataset), our method exhibits substantial improvement.
Remarkably, our method outperforms CoVeR~\cite{cover} with a significant margin (\textbf{74.6\%} \emph{vs.} 70.8\%), despite their adoption of a larger spatial resolution (448 \emph{vs.} 224), a pre-training dataset scale that is 1.5 times larger (3B \emph{vs.} 2B), and computational costs that are 7 times greater (17.6T \emph{vs.} 2.5T).

\noindent\textbf{Results on Kinetics-400}.
To verify the generalization ability of our method, we conduct experiments on the Kinetics-400 dataset, which does not rely as heavily on temporal modeling as the SS dataset.
Table~\ref{tab:k400_sota} presents our comparison with both CNN-based and transformer-based methods. Our ATM consistently outperforms CNN-based methods using both ResNet-50 and ResNet-101 backbones, showcasing its effectiveness in capturing temporal dynamics.
Moreover, we conducted a comprehensive comparison with recent transformer-based methods that utilize web-scale pre-training.  Notably, utilizing the same CLIP pre-trained backbones, our ATM achieved a substantial performance improvement over EVL~\cite{evl}, AIM~\cite{yang2023aim}, and ST-Adapter~\cite{st-adapter}, even when using fewer frames. 
It is worth mentioning that our method even performs comparably with more recent state-of-the-art approaches such as Text4Vis~\cite{text4vis,wu2023transferring} and BIKE~\cite{bike}, which integrate additional textual knowledge. Additionally, our method achieves superior performance compared to methods pre-trained with JFT-300M~\cite{JFT300M} or FLD-900M~\cite{yuan2021florence}, while requiring less computational cost.
Furthermore, with the significant scale-up of the pre-training data to 2 billion samples, our method attains an outstanding top accuracy of 89.4\%, solidifying its position as a state-of-the-art approach.

\section{Conclusions}
This paper provides an answer to the title question: The arithmetic operations between pairs of frame features are useful in temporal modeling by capturing expert knowledge.
To achieve this, we propose an Arithmetic Temporal Module (ATM) that performs addition, subtraction, multiplication, and division between pairs of extracted frame features, generating temporal cues. Then, we extract corresponding features from these cues back to the temporal-irrespective backbone. 
We comprehensively explore the instantiation of ATM on both CNN and Vision Transformer backbones and find that the combination of subtraction and multiplication yields the best performance on video recognition tasks.
Finally, we demonstrate that existing backbones plugged with ATM can achieve superior performance on a broad range of popular video benchmarks.




\section*{Acknowledgments}
This work was supported by the Australian Medical Research Future Fund MRFAI000085, CRC-P Smart Material Recovery Facility (SMRF) – Curby Soft Plastics, and CRC-P ARIA - Bionic Visual-Spatial Prosthesis for the Blind.

{\small
\bibliographystyle{ieee_fullname}
\bibliography{egbib}
}

\clearpage
\appendix
\section*{Appendix}
\setcounter{table}{0}
\setcounter{figure}{0}
\renewcommand{\thetable}{A.\arabic{table}}
\renewcommand{\thefigure}{A.\arabic{figure}}

In this appendix, we provide additional details as follows:
\S\ref{imp} contains more implementation details,
\S\ref{ablation} contains more ablation studies, \S\ref{results} contains additional results.


\section{Implementation Details}
\label{imp}

\subsection{Integration of ATM into Existing Backbones}\label{sec:insert}

\noindent\textbf{Integration of ATM into ResNet~\cite{resnet}.}    
As shown in Table~\ref{tab:resnet_arch}, we conduct experiments on two types of ResNet architectures:
ResNet with basic blocks, and ResNet with bottleneck blocks.
The ResNet basic block, which comprises two 3\x3 convolutional layers, is utilized in ResNet-18/34. In contrast, ResNet-50/101/152 includes the ResNet bottleneck block, consisting of two 1\x1 and one 3\x3 convolutional layers.
The ATM is integrated once after the last residual block of res$_3$, where the spatial resolution is 28\x28, in both two ResNet architectures. To optimize computational efficiency, we temporarily reduce the resolution to 14\x14 during the Context Spanning operation, followed by upsampling to 28\x28 during Domain Transformation.

\vspace{1mm}
\noindent\textbf{Integration of ATM into ViT~\cite{ViT}.} 
We seamlessly integrate the ATM into two representative ViT architectures: ViT-Base and ViT-Large, as provided by CLIP~\cite{CLIP}. These architectures are outlined in Table~\ref{vit}. In order to capture temporal cues for each anchor frame, we incorporate the ATM after the multi-head attention module of the $7$-th layer in ViT-B (which consists of 12 layers) and after the $14$-th layer in ViT-L (which consists of 24 layers).
To prepare the input for the ATM, we reshape the visual tokens within the ViT blocks, transforming them from one-dimensional spatial dimensions ($N = H\times W$) to two-dimensional spatial dimensions. As a result, we obtain $X^\texttt{IN} \in \mathbb{R}^{T\times C\times H\times W}$. We then apply the ATM directly to $X^\texttt{IN}$, similar to the way we apply it in ResNet.

\subsection{Training Hyperparameters}
All experiments are implemented in PyTorch. We use the configuration listed in Table~\ref{tb:imp} unless otherwise specified.

\newcommand{\blocks}[3]{\multirow{3}{*}{
    \(\left[\begin{array}{c}
        \text{1$\times$1$^\text{2}$, #2}\\[-.1em] 
        \text{1$\times$3$^\text{2}$, #2}\\[-.1em] 
        \text{1$\times$1$^\text{2}$, #1}
        \end{array}\right]\)$\times$#3}
}
\newcommand{\blockt}[3]{\multirow{3}{*}{\(\left[\begin{array}{c}\text{\underline{3$\times$1$^\text{2}$}, #2}\\[-.1em] \text{1$\times$3$^\text{2}$, #2}\\[-.1em] \text{1$\times$1$^\text{2}$, #1}\end{array}\right]\)$\times$#3}
}

\newcommand{\blocksb}[2]{\multirow{3}{*}{\(\left[\begin{array}{c}\text{1$\times$3$^\text{2}$, #1}\\[-.1em] \text{1$\times$3$^\text{2}$, #1}\end{array}\right]\)$\times$#2}
}

\newcommand{\blocktb}[2]{\multirow{3}{*}{\(\left[\begin{array}{c}\text{3$\times$3$^\text{2}$, #1}\\[-.1em] \text{3$\times$3$^\text{2}$, #1}\end{array}\right]\)$\times$#2}	
}	

{
\begin{table}[t]
	\centering
	\scalebox{0.86}{
    \setlength{\tabcolsep}{2.0pt}
		\begin{tabular}{c|c|c|c}
			stage & ResNet18 &  ResNet50 & output sizes\\
			\shline
			\multirow{1}{*}{raw clip} & - & - & 64\x224$^\text{2}$ \\
			\hline
			\multirow{2}{*}{data layer} & \multirow{2}{*}{stride 8, 1$^\text{2}$} & \multirow{2}{*}{stride 8, 1$^\text{2}$} &  \multirow{2}{*}{8\x$224^2$}   \\
			&  &  \\
			\hline
			\multirow{2}{*}{conv$_1$} & \multicolumn{1}{c|}{1\x7$^\text{2}$, {64}} & \multicolumn{1}{c|}{1\x7$^\text{2}$, {64}} &  \multirow{2}{*}{8\x$112^2$}    \\
			& stride 1, 2$^\text{2}$ & stride 1, 2$^\text{2}$  \\
			\hline
			\multirow{2}{*}{pool$_1$}  & \multicolumn{1}{c|}{1\x3$^\text{2}$ max} & \multicolumn{1}{c|}{1\x3$^\text{2}$ max} &  \multirow{2}{*}{8\x$56^2$}  \\
			& stride 1, 2$^\text{2}$ & stride 1, 2$^\text{2}$ & \\
			\hline
			\multirow{3}{*}{res$_2$} & \blocksb{{64}}{2} & \blocks{{256}}{{64}}{3} & \multirow{3}{*}{8\x$56^2$}   \\
			&  & \\
			&  & \\
			\hline
			\multirow{3}{*}{res$_3$} & \blocksb{{128}}{2} &  \blocks{{512}}{{128}}{4}  & \multirow{3}{*}{8\x$28^2$}   \\
			&  & \\
			&  & \\
			\hline
			\multirow{3}{*}{res$_4$} & \blocksb{{256}}{2} & \blocks{{1024}}{{256}}{6} &  \multirow{3}{*}{8\x$14^2$}   \\
			&  & \\
			&  & \\
			\hline
			\multirow{3}{*}{res$_5$} & \blocksb{{512}}{2} & \blocks{{2048}}{{512}}{3} &   \multirow{3}{*}{8\x$7^2$}  \\
			&  & \\
			&  & \\
			\hline
			\multicolumn{3}{c|}{global average pool, fc}  & \# classes \\
	\end{tabular}
	}
		\caption{\textbf{Two backbones of the ResNet}. 
		The dimensions of kernels are denoted by $\{ T \times S^2, C\}$ for temporal, spatial, and channel sizes.
		Strides are denoted as $\{$temporal stride, spatial stride$^2\}$.}
	\label{tab:resnet_arch}
\end{table}
}

\begin{table}[t]
\centering
\begin{tabular}{lcccc} \toprule
           & Embedding       & \multicolumn{3}{c}{Vision Transformer} \\
    Model   & dimension  & layers & width & heads \\ \midrule
    ViT-B/16 & 512  & 12 & 768 & 12  \\
    ViT-L/14 & 768  & 24 & 1024 & 16  \\
    \bottomrule
\end{tabular}
\caption{Two backbones of the ViT.}
\label{vit}
\end{table}

\begin{table*}
\centering
\begin{tabular}{lcccccc} \toprule
    \multirow{2}{*}{\textbf{Setting}} &  \multicolumn{2}{c}{\textbf{ResNet}} &  \multicolumn{4}{c}{\textbf{ViT}} \\ 
    & \textbf{SSV1/V2} & \textbf{K400} & \textbf{SSV1/V2} & \textbf{K400} & \textbf{ANet} & \textbf{Charades} \\ 
    \hline
    \rowcolor{gray!20}
    \multicolumn{7}{l}{\emph{Optimization}} \\
    Batch size & \multicolumn{2}{c}{64} & 256 & 256 & 256 & 256 \\
    Epochs & 60 & 150 &   20 ($B$), 15 ($L$) & 20 ($B$), 15 ($L$) & 15 & 15 \\
    Optimizer & \multicolumn{2}{c}{SGD} & \multicolumn{4}{c}{AdamW ($\beta_1=0.9, \beta_2=0.999$)} \\
    Initial LR & \multicolumn{2}{c}{0.01} & 7e-4 & 3e-4 ($B$), 2e-4 ($L$) & 2e-4 & 2e-4 \\
    Layer Decay & \multicolumn{2}{c}{-}  & 0.7 ($B$), 0.75 ($L$) & 0.65 ($B$), 0.7 ($L$) & 0.7 & 0.7 \\
    LR Schedule & \multicolumn{2}{c}{Multi-step, $\gamma=0.1$} & \multicolumn{4}{c}{Cosine}  \\ 
    LR Steps & [30,45,55] & [80,120,140] & \multicolumn{4}{c}{-} \\
    Weight Decay & \multicolumn{2}{c}{5e-4} & \multicolumn{4}{c}{0.05}  \\ 
    Linear Warm-Up &  \multicolumn{2}{c}{-} & \multicolumn{4}{c}{5 Epochs}\\
    Pre-training & \multicolumn{2}{c}{ImageNet-1K} & \multicolumn{4}{c}{WIT-400M} \\
    \hline
    \multicolumn{7}{l}{\cellcolor{baselinecolor}\emph{Augmentation}} \\
    Training Resize & \multicolumn{2}{c}{MultiScaleCrop} & \multicolumn{4}{c}{RandomSizedCrop} \\
    Rand Augment & \multicolumn{2}{c}{-} & rand-m7-n4-mstd0.5-inc1 & - & - & - \\
    Random Flip & \multicolumn{2}{c}{0.5} & \multicolumn{4}{c}{0.5} \\
    Repeated Sampling & \multicolumn{2}{c}{1} & \multicolumn{4}{c}{2} \\
    Label Smoothing & \multicolumn{2}{c}{-} & \multicolumn{4}{c}{0.1} \\
    GrayScale & \multicolumn{2}{c}{-} & - & 0.2 & 0.2 & 0.2 \\
    Mixup & \multicolumn{2}{c}{-} & \multicolumn{4}{c}{0.8} \\
    Cutmix & \multicolumn{2}{c}{-} & \multicolumn{4}{c}{1.0} \\
    \bottomrule
\end{tabular}
\vspace{-2mm}
\caption{Default training recipes. LR denotes the learning rate. $B$ represents ViT-Base and $L$ represents ViT-Large.}
\label{tb:imp}
\end{table*}



\section{Ablation Studies}
\label{ablation}
Here, we present additional ablation studies conducted on the Something-Something V1 dataset, specifically using ResNet-18 as the backbone and utilizing 8 frames.

\noindent\textbf{The Effect of Arithmetic Operations.} 
In our ATM, we employ context spanning to generate $L$ context frames for each frame, and then perform arithmetic operations between the features of the anchor frame and the $L$ context frames to extract temporal clues.
To evaluate the impact of arithmetic operations, we remove parameter-free arithmetic operations and directly use the features of context frames as temporal clues for the anchor frame. We find that this approach provides a basic level of temporal modeling.
Next, we incorporate the outputs of arithmetic operations between the anchor frame and context frames as our temporal clues. We observe that different arithmetic operations yield varying degrees of improvement in temporal modeling, as shown in Table~\ref{sup:atm}.

\begin{table}[h]
\centering
\begin{tabular}[t]{lc}
\toprule
Method &  Top-1  \\
\midrule
w/o ATM & 16.5\% \\ \midrule
ATM (Context features) & 40.6\% \\ \hdashline
ATM (\faPlus) &  41.1\% \\
ATM (\faMinus) & 43.6\% \\
ATM (\faTimes) & \textbf{46.5\%} \\
ATM (\faDivide) & 43.1\% \\
\bottomrule
\end{tabular}
\caption{The effect of arithmetic operations. Backbone: R18.}
\label{sup:atm}  
\end{table}

\begin{figure}[t]
\begin{center}
    \includegraphics[width=0.98\linewidth]{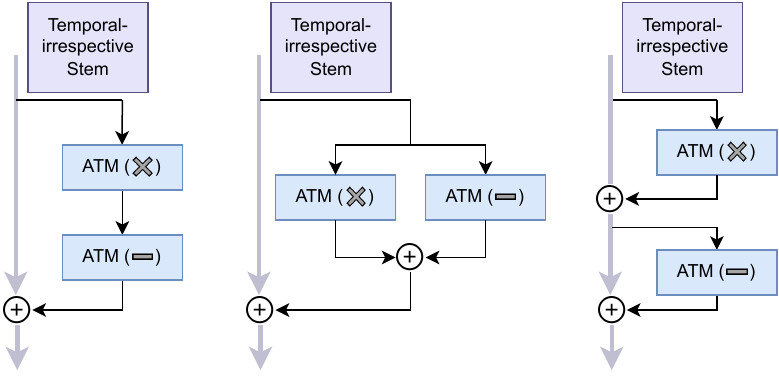} \\
    \begin{minipage}{0.27\linewidth}\subcaption{Cascade}\label{fig:a}\end{minipage}
    \begin{minipage}{0.38\linewidth}\subcaption{Parallel}\label{fig:b}\end{minipage}
    \begin{minipage}{0.32\linewidth}\subcaption{ATM-style}\label{fig:ATM}\end{minipage}
\end{center}
\caption{The different combinations of ATM(\faTimes) and ATM(\faMinus).}
\label{fig:comb}
\end{figure}

\noindent\textbf{Different Combinations of ATMs.}
In this part, we present several approaches for amalgamating ATMs (\eg, ATM (\faTimes) and ATM (\faMinus)). These approaches include (a) the cascade connection, (b) the parallel connection, and (c) the proposed ATM-style connection, as depicted in Figure~\ref{fig:comb}.
We present the experimental results in Table~\ref{sup:fusion}. The findings indicate that combining ATMs with either the cascade or parallel style leads to only marginal improvements over a single ATM (\faTimes). This result emphasizes the importance of domain transformation operations that transform signals to a temporal-irrespective stem.

\begin{table}[h]
  \centering
\begin{tabular}[t]{lc}
\toprule
Combinations &  Top-1  \\
\midrule
Single ATM (\faTimes) & 46.5\%  \\ 
Single ATM (\faMinus) & 43.6\% \\ \midrule
(a) Cascade & 46.6\%  \\ 
(b) Parallel & 46.9\%  \\
(c) ATM-style & \textbf{48.2\%}  \\
\bottomrule
\end{tabular}
\caption{Several combinations of ATMs. Backbone: R18.}
\label{sup:fusion}  
\end{table}


\begin{table*}[t]
    \centering
    \begin{tabular}{lccccccc}
    \toprule
      \multirow{2}{*}{Method}   & \multirow{2}{*}{Pretrain} & \multirow{2}*{Frame\x Crops\x Clip} & \multirow{2}{*}{GFLOPs} & \multicolumn{2}{c}{SSV1} & \multicolumn{2}{c}{SSV2} \\
      ~ & ~ & ~ & ~ & Top-1 & Top-5 & Top-1 & Top-5 \\ \midrule
      ATM ResNet50   & ImageNet-1K & 8\x1\x1 & 37\x1 & 53.9\% & 81.8\% & 65.5\% & 89.9\% \\
      ATM ResNet50   & ImageNet-1K & 8\x1\x2 & 37\x2 & 54.7\% & 82.6\% & 66.1\% & 90.2\% \\
      ATM ResNet50   & ImageNet-1K & 16\x1\x1 & 74\x1 & 56.3\% & 83.4\% & 67.4\% & 91.1\% \\
      ATM ResNet50   & ImageNet-1K & 16\x1\x2 & 74\x2 & 56.7\% & 83.6\% & 67.6\% & 91.2\% \\
      ATM ResNet50   & ImageNet-1K & 32\x1\x1 & 148\x1 & 57.1\% & 84.0\%  & 68.4\% & 91.5\% \\
      ATM ResNet50   & ImageNet-1K & 32\x1\x2 & 148\x2 & 57.2\% & 84.3\% & 68.5\% & 91.6\% \\
      ATM ResNet50   & ImageNet-1K & (8+16)\x1\x1 & 111\x1 & 57.6\% & 84.4\% & 68.3\% & 91.6\% \\
      ATM ResNet50   & ImageNet-1K & (8+16)\x1\x2 & 111\x2 & 57.8\% & 84.8\% & 68.7\% & 91.7\% \\
      ATM ResNet50   & ImageNet-1K & (8+16+32)\x1\x1 & 259\x1 & 59.1\% & 85.7\% & 69.7\% & 92.4\% \\
      \hdashline
      ATM ResNet101   & ImageNet-1K & 8\x1\x1 & 67\x1 & 54.9\% & 82.3\% & 66.4\% & 90.3\% \\
      ATM ResNet101   & ImageNet-1K & 8\x1\x2 & 67\x2 & 55.8\% & 82.9\% & 66.9\% & 90.7\% \\
      ATM ResNet101   & ImageNet-1K & 16\x1\x1 & 134\x1 & 57.2\% & 84.1\% & 68.2\% & 91.5\% \\
      ATM ResNet101   & ImageNet-1K & 16\x1\x2 & 134\x2 & 57.4\% & 84.4\% & 68.6\% & 91.6\% \\
      ATM ResNet101   & ImageNet-1K & 32\x1\x1 & 268\x1 & 57.9\% & 84.3\% & 69.3\% & 92.0\% \\
      ATM ResNet101   & ImageNet-1K & (8+16)\x1\x1 & 201\x1 & 58.6\% & 84.9\% & 69.4\% & 92.1\% \\
      ATM ResNet101   & ImageNet-1K & (8+16)\x1\x2 & 201\x2 & 58.9\% & 85.1\% & 69.6\% & 92.3\% \\
      ATM ResNet101   & ImageNet-1K & (8+16+32)\x1\x1 & 469\x1 & 60.0\% & 86.1\% & 70.8\% & 92.9\% \\ \midrule
      ATM ViT-B/16   & WIT-400M & 8\x1\x1 & 99\x1 & 58.1\% & 84.7\% & 69.4\% & 92.2\% \\
      ATM ViT-B/16   & WIT-400M & 8\x3\x2 & 99\x6 & 58.8\% & 85.4\% & 70.5\% & 92.7\% \\
      ATM ViT-B/16   & WIT-400M & 16\x1\x1 & 198\x1 & 59.5\% & 86.1\% & 70.9\% & 92.8\% \\
      ATM ViT-B/16   & WIT-400M & 16\x3\x2 & 198\x6 & 60.6\% & 86.5\% & 71.5\% & 93.0\% \\
      ATM ViT-B/16   & WIT-400M & 32\x1\x1 & 378\x1 & 60.9\% & 85.9\% & 71.6\% & 93.2\% \\  
      ATM ViT-B/16   & WIT-400M & 32\x3\x2 & 378\x6 & 61.5\% & 86.2\% & 71.9\% & 93.3\% \\  \hdashline
      ATM ViT-L/14   & WIT-400M & 8\x1\x1 & 421\x1 & 61.9\% & 87.0\% & 71.3\% & 93.1\% \\
      ATM ViT-L/14   & WIT-400M & 8\x3\x2 & 421\x6 & 62.8\% & 87.6\% & 72.1\% & 93.6\% \\
      ATM ViT-L/14   & WIT-400M & 16\x1\x1 & 842\x1 & 63.3\% & 87.5\% & 73.1\% & 93.5\% \\
      ATM ViT-L/14   & WIT-400M & 16\x3\x1 & 842\x3 & 63.7\% & 88.0\% & 73.2\% & 93.7\% \\
      ATM ViT-L/14   & WIT-400M & 16\x3\x2 & 842\x6 & 64.0\% & 88.0\% & 73.5\% & 93.7\% \\
      ATM ViT-L/14   & Merged-2B & 8\x3\x2 & 421\x6 & 64.9\% & 88.9\% & 73.7\% & 94.1\% \\
      ATM ViT-L/14   & Merged-2B & 16\x3\x2 & 842\x6 & 65.6\% & 88.6\% & 74.6\% & 94.4\% \\
      \bottomrule
    \end{tabular}
    \caption{More results on Something-Something V1 \& V2.}
    \label{tab:ss}
\end{table*}

\begin{table*}[t]
    \centering
    \begin{tabular}{lccccc}
    \toprule
      Method   & Pretrain & Frame\x Crops\x Clip & GFLOPs & Top-1 & Top-5 \\ \midrule
      ATM ResNet50   & ImageNet-1K & 8\x3\x10 & 37\x30 & 77.0\% & 92.9\% \\
      ATM ResNet50   & ImageNet-1K & 16\x3\x10 & 74\x30 & 77.6\% & 93.2\% \\ 
      ATM ResNet101   & ImageNet-1K & 8\x3\x10 & 67\x30 & 78.0\% & 93.5\% \\
      ATM ResNet101   & ImageNet-1K & 16\x3\x10 & 134\x30 & 78.8\% & 93.7\% \\ 
      ATM ResNet152   & ImageNet-1K & 16\x3\x10 & 191\x30 & 79.4\% & 93.7\% \\
      ATM R101+R152   & ImageNet-1K & (16+16)\x3\x10 & 326\x30 & 80.5\% & 94.4\% \\ \midrule
      ATM ViT-B/16   & WIT-400M & 8\x3\x4 & 99\x12 & 84.1\% & 96.3\% \\
      ATM ViT-L/14   & WIT-400M & 8\x3\x4 & 421\x12 & 87.3\% & 97.4\% \\
      ATM ViT-L/14   & WIT-400M & 32\x3\x4 & 1684\x12 & 88.0\% & 97.6\% \\
      ATM ViT-L/14 (336$\uparrow$)   & WIT-400M & 32\x3\x4 & 3784\x12 & 88.2\% & 97.9\% \\
      ATM ViT-L/14   & Merged-2B & 8\x3\x4 & 421\x12 & 88.0\% & 97.6\% \\
      ATM ViT-L/14 (336$\uparrow$)   & Merged-2B & 8\x3\x4 & 946\x12 & 88.9\% & 97.8\% \\
      ATM ViT-L/14 (336$\uparrow$)   & Merged-2B & 32\x3\x4 & 3784\x12 & 89.4\% & 98.3\% \\
      \bottomrule
    \end{tabular}
    \caption{More results on Kinetics-400.}
    \label{tab:k400}
\end{table*}

\section{Additional Results}
\label{results}

\textbf{More Results on Kinetics-400, Something-Something V1 \& V2.} For readers' reference, we present our results with various views in Table~\ref{tab:k400} and Table~\ref{tab:ss}.

\textbf{Results on ActivityNet.}
To demonstrate the generalization ability of our method, we evaluate its performance on the widely-used untrimmed video benchmark, ActivityNet-v1.3~\cite{caba2015activitynet}. This dataset consists of 19,994 videos ranging from 5 to 10 minutes in length, covering 200 activity categories. We fine-tune the CLIP pre-trained ViT-L backbone with 16 frames on this dataset and report the top-1 accuracy and mean average precision (mAP) using official evaluation metrics. As shown in Table~\ref{tab:anet_sota}, our method outperforms recent works, achieving an mAP accuracy of 94.7\%.

\textbf{Results on Charades.} We also conduct experiments on the multi-label video recognition task using the Charades dataset~\cite{charades}. This dataset consists of over 10,000 short video clips covering 157 action categories. We trained the CLIP pre-trained ViT-L backbone for this task and evaluated the results using the Mean Average Precision (mAP) metric. Table~\ref{t:charades} illustrates the effectiveness of our method in multi-label video classification.

\begin{table}[h]
\centering
\begin{tabular}{lcc} \toprule
  Method   & Top-1 & mAP \\ \midrule
   ListenToLook \cite{gao2020listen}  & - & 89.9 \\
   MARL \cite{wu2019multi} & 85.7 & 90.1 \\ 
   DSANet \cite{dsanet} & - & 90.5 \\ 
   TSQNet \cite{tsqnet} & 88.7 & 93.7 \\ 
   \midrule
   Ours ViT-L & \textbf{90.2} & \textbf{94.7} \\ 
   \bottomrule
\end{tabular}
    \caption{Comparisons with previous works on ActivityNet.}
    \label{tab:anet_sota}
\end{table}

\begin{table}[h]
\centering
\begin{tabular}{lcc}
\toprule
\textbf{Method}   & \textbf{Frames}  & \textbf{mAP} \\ \midrule
MultiScale TRN \cite{trn} & - &  25.2\% \\
STM \cite{stm} & 16 &  35.3\% \\
Nonlocal \cite{nonlocal} & - &  37.5\% \\
SlowFast R50 \cite{slowfast} & 8+32 &  38.0\% \\
SlowFast R101 \cite{slowfast} & 16+64 &  42.5\% \\
LFB+NL \cite{lfb}  & 32 &  42.5\% \\
X3D-XL (312$\uparrow$) \cite{feichtenhofer2020x3d}  & 16 &  43.4\% \\
ActionCLIP \cite{wang2021actionclip} & 32 & 44.3\% \\
\midrule
Ours ViT-L & 16 & \textbf{48.5\%} \\
\bottomrule
\end{tabular}
\caption{Comparison with previous works on \textbf{Multi-Label} video dataset Charades.}
\label{t:charades}
\end{table}

\end{document}